\documentclass{article}

\usepackage{arxiv}
\usepackage[utf8]{inputenc}
\usepackage[T1]{fontenc}
\usepackage{hyperref}
\usepackage{url}
\usepackage{booktabs}
\usepackage{amsmath,amsfonts,amssymb}
\usepackage{mathtools}
\usepackage{microtype}
\usepackage{graphicx}
\usepackage[round]{natbib}
\usepackage{doi}
\usepackage{float}
\usepackage{xcolor}
\usepackage{textcomp}
\usepackage[normalem]{ulem}

\newcommand{\enluminure}[2]{#1#2}
\newcommand{\acks}[1]{\section*{Acknowledgments}#1}

\newcommand{\data}[1]{\section*{Data Availability}#1}

\title{When Do Histology Foundation Models Need Geometric Test-Time Augmentation?}

\newcommand{\orcidlink}[2]{\href{https://orcid.org/#1}{\IfFileExists{latex/orcid.png}{\includegraphics[scale=0.008]{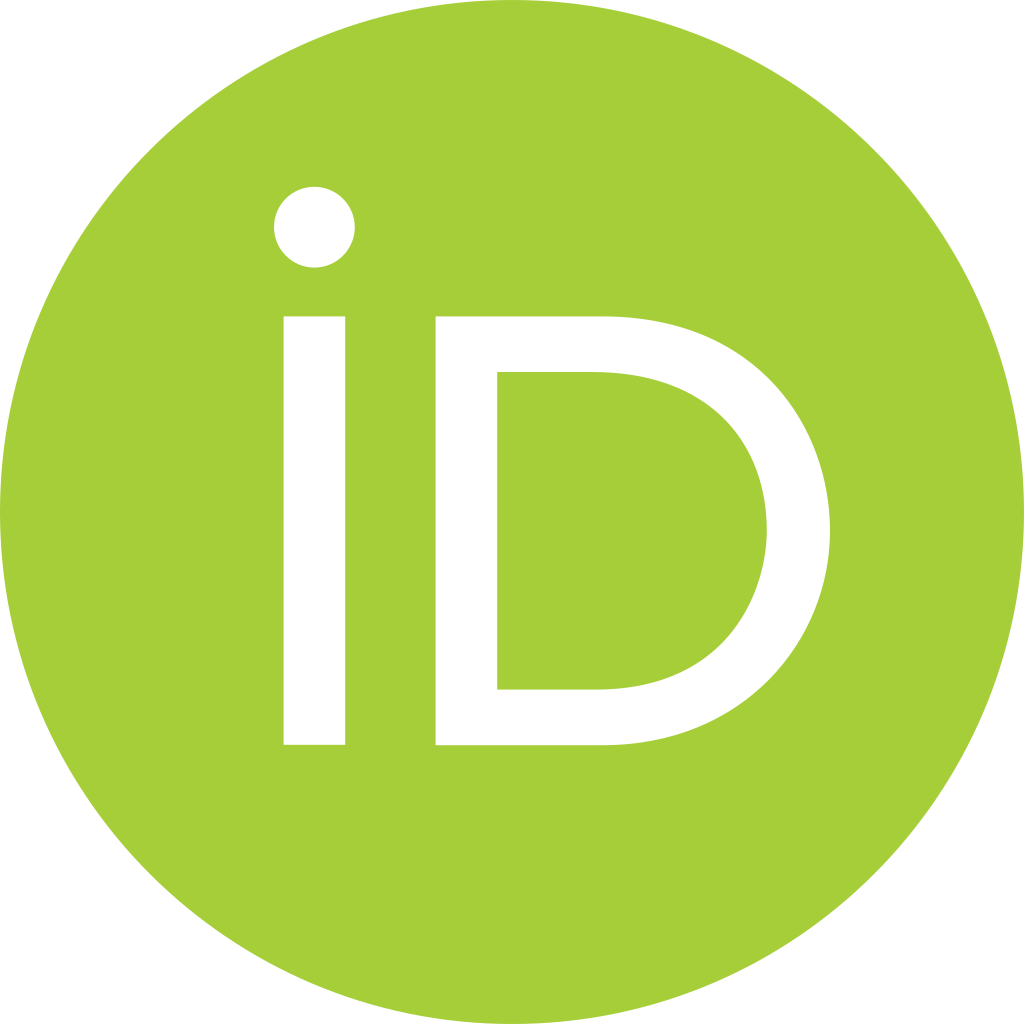}\hspace{1mm}}{}#2}}

\author{
	\orcidlink{0009-0001-1319-3097}{Lars Chen}$^{1,2}$\thanks{Corresponding author: \texttt{lars.chen@nyulangone.edu}} \quad
	\orcidlink{0000-0002-7864-7053}{Valentino F.~Foit}$^{3}$ \quad
	Dorry Segev$^{1}$ \quad
	\orcidlink{0000-0002-9923-7408}{Michal Mankowski}$^{1}$ \\[1ex]
	$^{1}$Department of Surgery, New York University Langone Health, USA \\
	$^{2}$Hertie Institute for AI in Brain Health, University of T\"ubingen, Germany \\
	$^{3}$Department of Physics, New York University, USA \\
}

\date{}   
\renewcommand{\shorttitle}{TTA for Histology FMs}
\renewcommand{\undertitle}{}   
\renewcommand{\headeright}{}   
\fancypagestyle{plain}{\fancyhf{}\fancyfoot[C]{\thepage}\fancyfoot[L]{\small\itshape Preprint. Under review.}} 

\hypersetup{
pdftitle={When Do Histology Foundation Models Need Geometric Test-Time Augmentation?},
pdfsubject={cs.CV, eess.IV, q-bio.QM},
pdfauthor={Lars Chen, Valentino F. Foit, Dorry Segev, Michal Mankowski},
pdfkeywords={Histopathology, Test-Time Augmentation, Foundation Models},
hidelinks,
}

\begin{document}
\maketitle
\thispagestyle{plain} 
\vspace{-2.5em} 

\begin{abstract}
Test-time augmentation (TTA) aggregates model predictions across multiple transformed views of an input during inference. Because histology images lack a canonical orientation, $D_4$ TTA---pooling across the 8 rotations and reflections of a square---is a standard default in computational pathology. However, the rise of self-supervised histology foundation models (FMs) with inherent rotational invariance calls its continued utility into question. We benchmark 9 histology FMs and 6 general-purpose models across four patch-level and two whole-slide image (WSI) datasets. We find that $D_4$ TTA remains a reliable post hoc improvement, but benefits diminish from $+1.98$ percentage points (pp) for general-purpose backbones to $+0.50$ pp for histology FMs on TCGA-UT, and shrink further for the largest models. To eliminate computational overhead, our experiments demonstrate that entropy routing---applying TTA selectively to uncertain predictions---recovers most of the gain while augmenting only ${\sim}13\%$ of samples. Finally, we show that patch-level TTA improvements do not transfer to slide-level multiple instance learning (MIL).
Code:~\url{https://github.com/lars-chen/histology-tta}
\end{abstract}

\vspace{1.5em}

\begin{figure*}[!tb]
    \centering
    \includegraphics[width=1.0\linewidth]{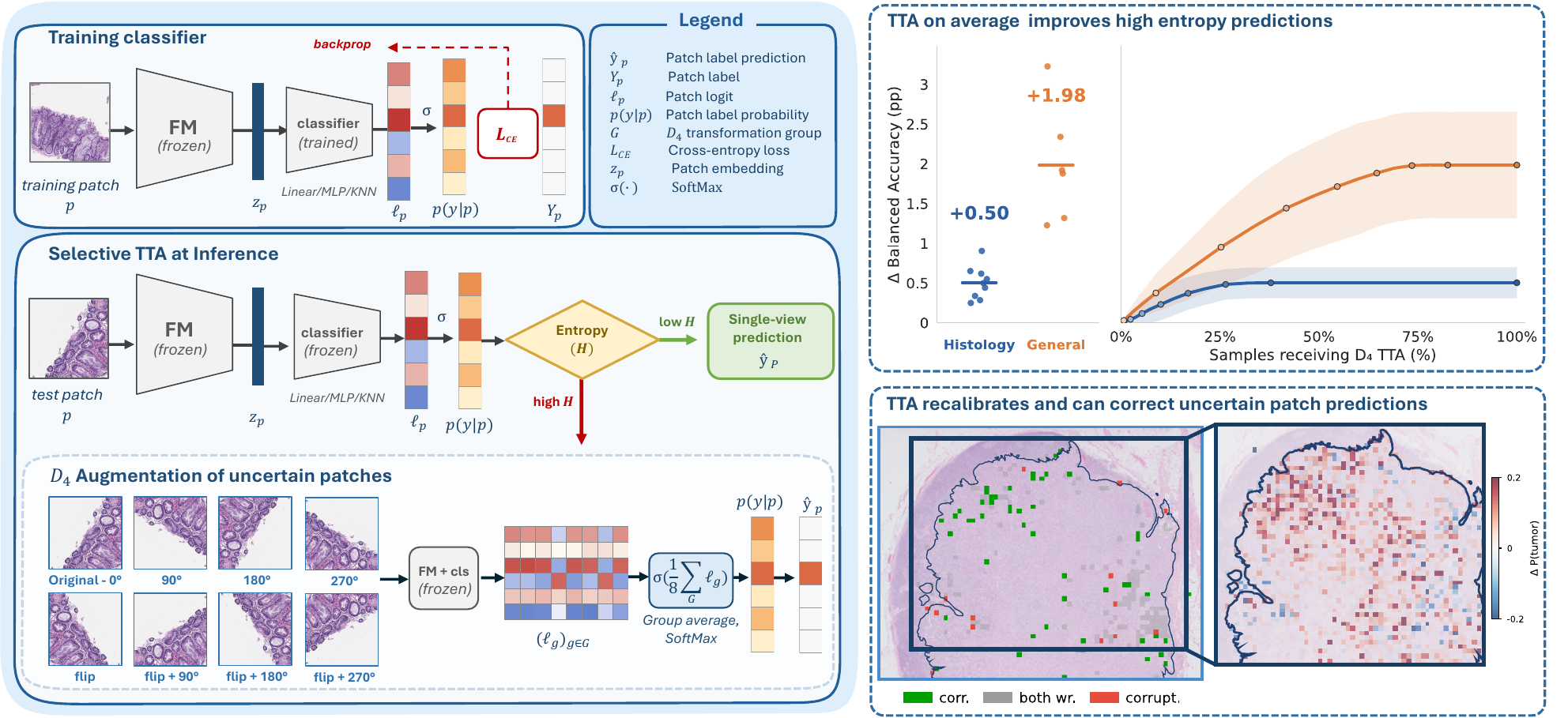}
    \caption{$D_4$ test-time augmentation with selective routing improves histology foundation model classification. (Left) A lightweight classifier is trained using FM-extracted patch embeddings $z_p$. At inference, patch prediction entropy $H$ gates augmentation: low-entropy patches receive a single-view prediction $\hat{y}_p$; high-entropy patches are passed through the $D_4$ symmetry group and predictions are averaged across views. (Top right) $D_4$ TTA improves balanced accuracy across both histology FMs and general-purpose models. Dots show per-model gains, on TCGA-UT, with the mean bar indicating +0.5 pp and +1.98 pp respectively. Applying TTA selectively with entropy thresholds (curves, averaged across models) recovers most of this gain while augmenting only a fraction of patches. (Bottom right) Applied to all patches in an exemplar Camelyon17 slide~\citep{bandi2018camelyon17}, full TTA corrects predictions from UNI embeddings. Green patches indicate corrected and blue line shows tumor boundary. Zoomed image shows the overall increase in tumor probability within boundary.}
    \label{fig:graphical_overview}
\end{figure*}

\section{Introduction}

\enluminure{H}{istopathology} image analysis increasingly relies on foundation models (FMs): large vision transformers pretrained on millions of hematoxylin and eosin (H\&E) tissue patches by self-supervised learning~\citep{chen2024uni,vorontsov2024virchow,filiot2023phikon,xu2024gigapath,hoptimus2024}. The strong downstream performance of FMs has made them the standard feature extractor in computational pathology. 

Because tissue patches have no canonical orientation, evaluating all eight dihedral-group views at inference and aggregating predictions is a principled inference-time strategy for enforcing rotational and reflection invariance without retraining. Given recent advances in pretraining scale and objectives, however, whether test-time augmentation (TTA) still yields meaningful gains for FM-based classifiers remains an open question. We provide a systematic evaluation of this strategy across modern FMs commonly deployed in computational pathology pipelines~\citep{lu2021clam,zhao2025smmile}. Figure~\ref{fig:graphical_overview} provides a graphical overview of our approach, and Table~\ref{tab:datasets} summarizes the datasets used in our experiments.

Tissue patches are approximately symmetric under the dihedral group $D_4$, i.e., the eight isometries of a square, comprising four rotations (0\textdegree, 90\textdegree, 180\textdegree, 270\textdegree) and their respective reflections. An ideal classifier should therefore be invariant to these transforms. This can be achieved in two ways: training-time augmentation applies random $D_4$ transforms during training, encouraging the model to learn invariant representations; instead, TTA averages predictions across all eight views at inference, approximating invariance post hoc without modifying the trained model.

Group-averaged TTA corresponds to applying the Reynolds operator~\citep{elesedy2021provably}. For a group of transformations $G$ acting on input space $\mathcal{X}$ and a function $f: \mathcal{X} \to \mathbb{R}^C$,
\begin{equation}
\hat{f}(x) = \frac{1}{|G|} \sum_{g \in G} f(g \cdot x)
\label{eq:reynolds}
\end{equation}
produces the unique $G$-invariant function closest to $f$ in $L^2$ norm when the data are $G$-invariant, as histology patches are, justifying averaging as the optimal strategy to enforce invariance. In practice, however, averaging can correct some predictions while corrupting others. When $f$ is already $G$-invariant, the Reynolds operator reduces to the identity and TTA provides no benefit.

Hence, we investigate to what extent modern histology FMs are $D_4$ invariant, and under what conditions TTA provides meaningful gains. We address this through a large-scale TTA evaluation in computational pathology, spanning:
\begin{itemize}
    \item \textbf{16 models} from three families: 9 histology-specialized FMs, 6 general-purpose vision models, and 1 $D_4$-equivariant reference model (Table~\ref{tab:models})
    \item \textbf{4 patch datasets} of varying difficulty: TCGA-UT (31 classes)~\citep{komura2022universal}, NCT-CRC-HE-100K and NCT-CRC-HE-NoNorm (9 classes each)~\citep{kather2019predicting}, and MHIST (2 classes)~\citep{wei2021petri}
    \item \textbf{2 whole-slide image (WSI) datasets} for slide-level TTA analysis: Camelyon17 (lymph node sections)~\citep{bandi2018camelyon17}, and PANDA (prostate biopsy samples)~\citep{bulten2022artificial}, covering both patch-level and weakly supervised multiple instance learning (MIL) classification
\end{itemize}
\begin{table}[t]
    \centering
    \caption{Models evaluated in this study. Histology FMs are pretrained on histopathology data via self-supervised learning. General-purpose models are pretrained on ImageNet. The equivariant model (D4WRN) has architectural $D_4$ symmetry guarantees. Number of parameters and training patches are reported in millions (M). $^\dagger$TCGA-UT slides may appear in self-supervised pretraining, however without labels (Section~\ref{sec:tta_improves}).}
    \label{tab:models}
    \vspace{3pt}
    \begin{tabular}{@{}llrrl@{}}
        \toprule
        Model & Type & Params & Patches & Arch. \\
        \midrule
        CTransPath$^\dagger$      & Histo      & 28   & 15    & Swin-T \\
        Phikon$^\dagger$      & Histo      & 86   & 40    & ViT-B/16 \\
        Phikon-v2$^\dagger$   & Histo      & 300  & 456   & ViT-L/16 \\
        UNI         & Histo      & 300  & 100   & ViT-L/16 \\
        UNI-v2      & Histo      & 681  & 200   & ViT-H/14 \\
        Virchow     & Histo      & 632  & 2000  & ViT-H/14 \\
        Virchow-v2  & Histo      & 632  & 2000  & ViT-H/14 \\
        GigaPath    & Histo      & 1100 & 1400  & ViT-g/16 \\
        H-Optimus-1 & Histo      & 1100 & $>$1000 & ViT-g/14 \\
        \midrule
        ResNet-18    & General    & 12   & --    & ResNet \\
        ResNet-50    & General    & 26   & --    & ResNet \\
        ConvNeXtV2-T & General    & 28   & --    & ConvNeXt \\
        ConvNeXtV2-B & General    & 89   & --    & ConvNeXt \\
        DINOv2-S     & General    & 22   & --    & ViT-S/14 \\
        DINOv2-B     & General    & 86   & --    & ViT-B/14 \\
        \midrule
        D4WRN        & Equivariant & 12  & --    & Wide ResNet \\
        \bottomrule
    \end{tabular}
\end{table}

Our key finding is that $D_4$ TTA is a reliable, modest post hoc improvement for patch classification, even with strong histology FMs. On TCGA-UT, histology FMs gain $+0.5$ pp in balanced accuracy on average, while general-purpose backbones benefit $+2$ pp. TTA improves predictions over all four patch benchmarks, but the average gains are smaller in higher baseline tasks (see Figure~\ref{fig:tta_overview} and Table~\ref{tab:full_results}).

At inference time, the full $G=D_4$ TTA increases the computational cost by a factor of $|G|=8$. For histology FMs, where the average gain is small, this cost could be hard to justify in routine clinical or research use. By contrast, general-purpose backbones exhibit greater view-to-view disagreement and weaker class separability, so TTA yields larger average improvements and can be worthwhile. We therefore evaluate entropy-based selective TTA~\citep{son2023efficient} to determine how much overhead can be avoided while retaining most of the performance benefit across models with different baseline degrees of rotational invariance. 

Finally, we extend the analysis beyond independent tiles to WSIs. Because these gigapixel-scale clinical slides require aggregating thousands of patches via MIL, we test whether localized tile-level corrections survive this process to improve overall slide-level performance.

\section{Related Works}
\label{sec:related}

\subsection{Test-Time Augmentation}

TTA has a long history in computer vision, from early ensemble methods~\citep{krizhevsky2012imagenet} to systematic studies of augmentation policies and aggregation~\citep{shanmugam2021better,kimTTAsurvey2020}. In biomedical imaging, it is routinely applied in histopathology~\citep{liu2017detecting}, microscopy cell segmentation~\citep{moshkov2020test}, and dermatology~\citep{perez2018data}. Most prior work studies whether TTA helps a particular model on a particular task, without systematically evaluating how its benefit varies across modern backbone families (notably transformer-based models) or across different classifier heads and pretraining strategies.

\citet{elesedy2021provably} prove that averaging a predictor over a symmetry group strictly improves generalization when the target is invariant. Beyond computer vision, \citet{foit2025group} recently demonstrated the efficacy of group-averaged TTA for autoregressive prediction in physics.

\citet{shanmugam2021better} find that TTA, while improving accuracy overall, also turns some correct predictions into incorrect ones, and that learned aggregation outperforms simple averaging. TTA also multiplies inference cost, so \citet{son2023efficient} propose selective TTA that routes uncertain samples to additional augmentations. However, it remains unclear how $D_4$ TTA and selective routing behave across modern histology FMs versus general-purpose backbones.

\subsection{Foundation Models for Histopathology}
\label{sec:related_fm}

The past two years have seen an explosion of histology FMs pretrained by self-supervised learning on large histopathology corpora (Table~\ref{tab:models}): CTransPath~\citep{wang2022ctranspath}, Phikon~\citep{filiot2023phikon} and Phikon-v2~\citep{filiot2024phikon2}, UNI~\citep{chen2024uni} and UNI-v2, Virchow~\citep{vorontsov2024virchow} and Virchow-v2~\citep{zimmermann2024virchow2}, GigaPath~\citep{xu2024gigapath} and H-Optimus-1~\citep{hoptimus2024}. They range from 28M to 1.1B parameters and from tens of millions to billions of training tiles, and most are pretrained with DINOv2~\citep{oquab2023dinov2}; Phikon uses iBOT~\citep{zhou2021ibot} and CTransPath a contrastive objective.

A common feature of these models' pretraining is strong stochastic augmentation, including geometric ones such as crops, flips and rotations, used within self-supervised learning objectives such as DINO and iBOT~\citep{caron2021dino,oquab2023dinov2,zhou2021ibot}. By explicitly matching model outputs across differently augmented views of the same patch~\citep{caron2021dino,zhou2021ibot}, these objectives encourage invariance to the augmentation set, promoting $D_4$-stable representations when $D_4$ transforms are included. Along with the orientation diversity of histopathology data, these objectives would be expected to yield approximately $D_4$-invariant representations. We find instead that histology FM embeddings are about as $D_4$-invariant as those of general-purpose models (Section~\ref{sec:orbit}); their prediction stability arises from classification margin.

In addition to histology-specialized FMs, computational pathology pipelines frequently rely on general-purpose FMs (e.g., ImageNet-pretrained CNNs and self-supervised transformers such as DINOv2) as tile encoders and baselines due to their accessibility and strong performance~\citep{lu2021clam,zhao2025smmile,oquab2023dinov2}. Because these models are not trained on histology with the same orientation diversity as histology FMs, they provide a useful contrast case in our study.

\subsection{Equivariant Neural Networks}

Group-equivariant convolutional networks~\citep{cohen2016group,weiler2019general} provide architectural guarantees of equivariance to specified symmetry groups. For the $D_4$ group, architectures like E2-CNNs~\citep{weiler2019general} and $p4m$-CNNs~\citep{cohen2016group} ensure that feature maps transform predictably under rotations and reflections. Rotation-equivariant architectures have shown benefits in microscopy and histopathology~\citep{chidester2019rotation,veeling2018rotation,lafarge2021roto}. We construct D4WRN, a Wide ResNet with a $D_4$-equivariant backbone and an invariant readout, as a reference point: its predictions are $D_4$-invariant by construction, so its TTA benefit should be near zero.

Prior work establishes TTA as a generally useful ensembling technique and motivates $D_4$ symmetry for histology. \citet{elphick2024rotation} found that FMs pretrained with rotation augmentation have tighter rotation orbits in embedding space, but did not examine downstream predictions or TTA. We show that orbit size barely predicts whether a prediction changes under $D_4$; what matters is the relative distance to the decision boundary (Section~\ref{sec:orbit}).

\begin{figure*}[!tb]
    \centering
    \includegraphics[width=1.0\linewidth]{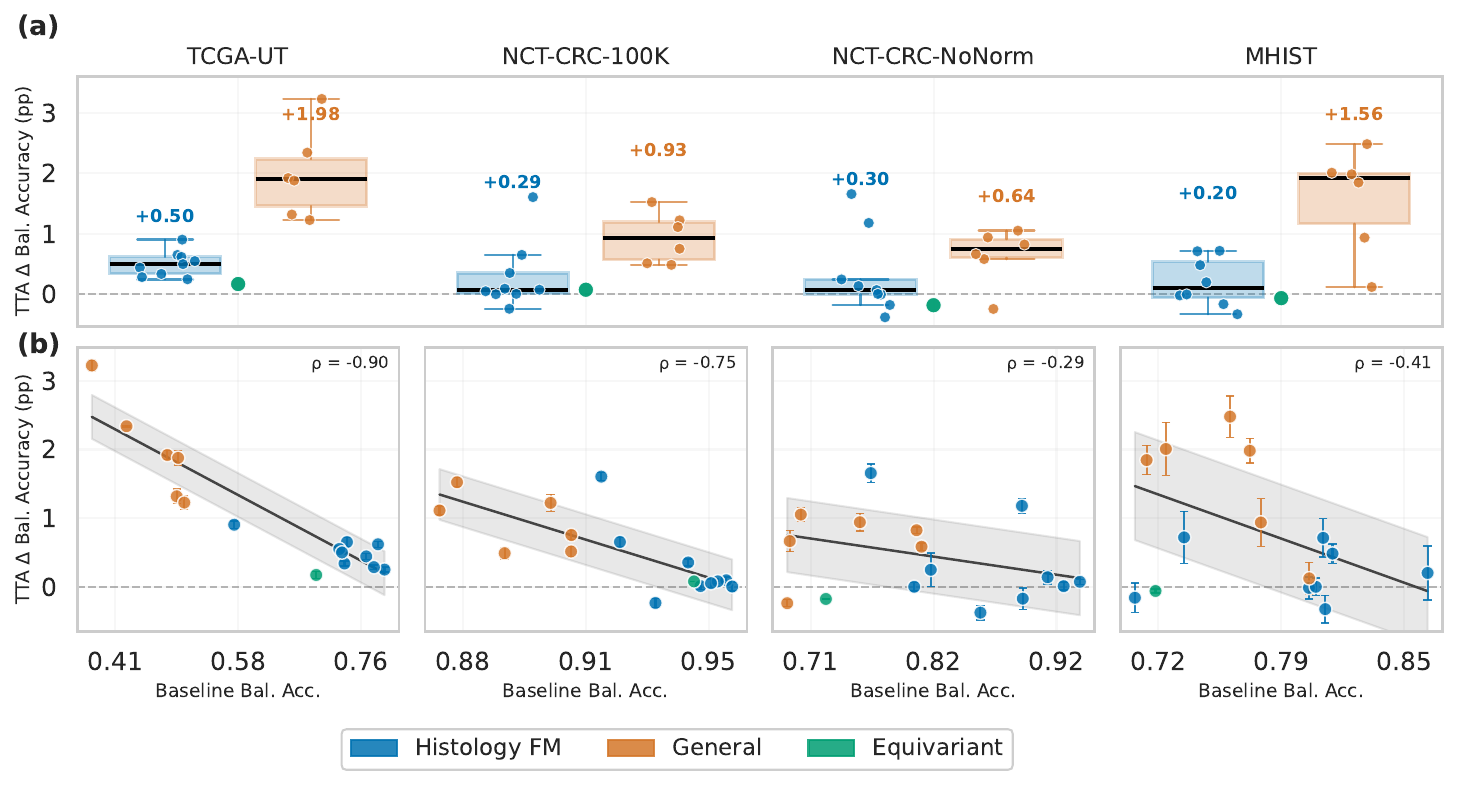}
    \caption{$D_4$ TTA consistently improves histopathology classifiers, with stronger gains for weaker baselines.
(a) Distribution of $\Delta_{\text{TTA}}$ in balanced accuracy (pp) across models for each dataset, split by model family (histology FMs vs. general-purpose backbones). Points show individual models; boxplots summarize the distribution; D4WRN (equivariant reference) exhibits near-zero gain by construction.
(b) Relationship between baseline balanced accuracy and $\Delta_{\text{TTA}}$ for each dataset: weaker baselines tend to benefit more from TTA. }
    \label{fig:tta_overview}
\end{figure*}

\begin{figure*}[!tb]
    \centering
    \includegraphics[width=1.0\linewidth]{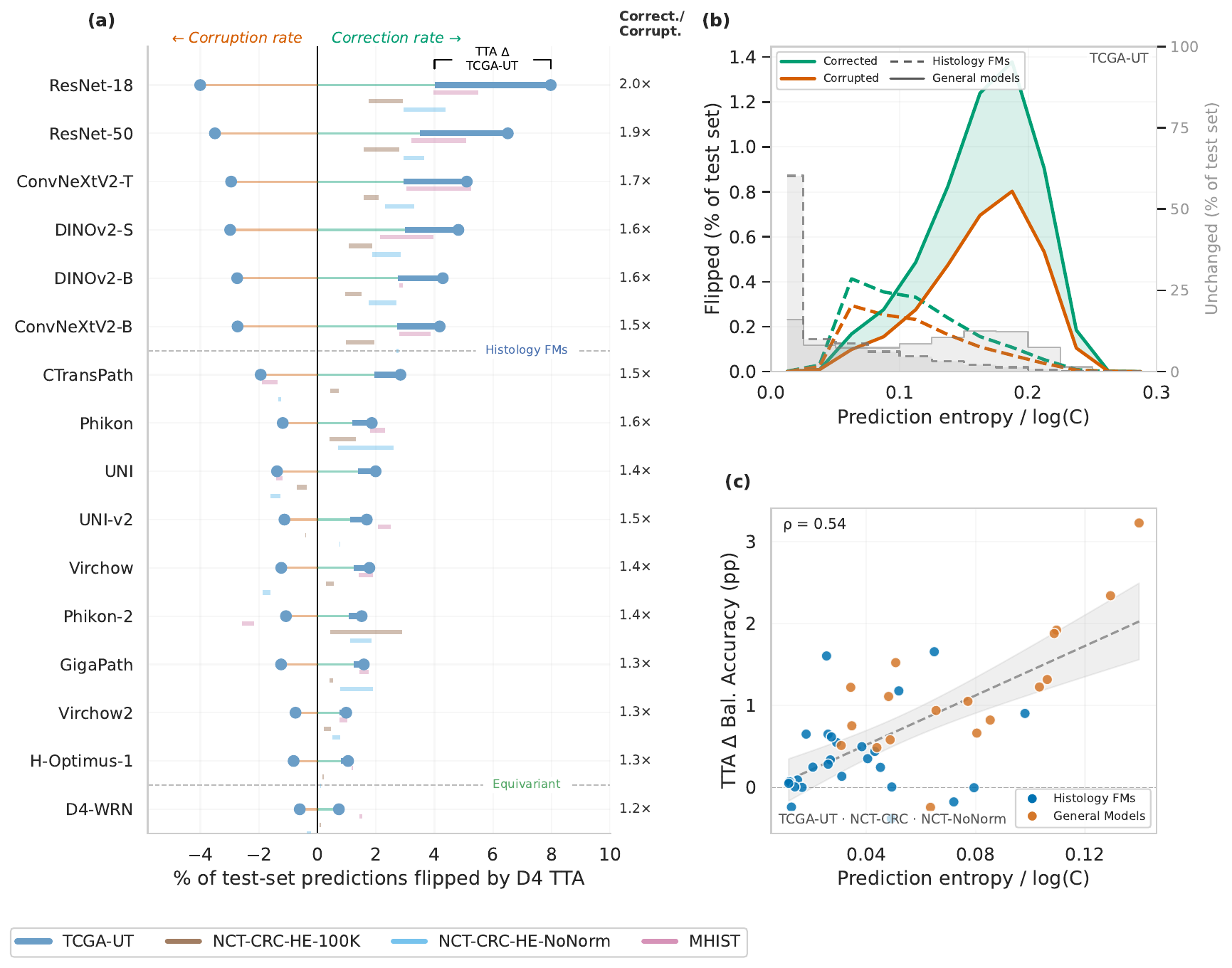}
    \caption{Prediction entropy explains $D_4$ TTA outcomes.
(a) Per model (TCGA-UT; other datasets as offset segments), predictions flipped by $D_4$ TTA split into \emph{corrections} and \emph{corruptions}, sorted by $\Delta_{\text{TTA}}$. The thick segment is the net effect; the right margin gives the correction/corruption ratio. Models with small $\Delta_{\text{TTA}}$ flip few predictions.
(b) On TCGA-UT, binned by baseline normalized entropy: flips and net gains concentrate in high-entropy predictions, where corrections outpace corruptions.
(c) Model-level association (MHIST excluded). Mean entropy is moderately associated with $\Delta_{\text{TTA}}$ (Spearman $\rho = 0.54$, $p = 10^{-4}$); within datasets $\rho = 0.89$ (TCGA-UT), $0.71$ (NCT-CRC-HE-100K) and $0.25$ (NCT-CRC-HE-NoNorm, where stain variation dominates; Section~\ref{sec:stain}). This trend is descriptive; routing rests on the per-patch results in (b) and Table~\ref{tab:flip_predictors}, and is evaluated in Section~\ref{sec:selective}.}
    \label{fig:entropy}
\end{figure*}

\section{Methods}
\label{sec:methods}

\subsection{Models}

We evaluate 16 models spanning three categories (Table~\ref{tab:models}).

\textbf{Histology foundation models} (9 models) are used with frozen backbones and a trained linear classification head, following the standard linear probing protocol. This isolates the quality of the pretrained representations without confounding from downstream finetuning.

\textbf{General-purpose models} (6 models) pretrained on ImageNet are evaluated in both frozen (linear probe) and finetuned configurations to assess how downstream adaptation affects TTA benefit.

\textbf{Equivariant reference} (1 model): a $D_4$-equivariant Wide ResNet, D4WRN~\citep{zagoruyko2016wide,weiler2019general} trained from scratch, providing an architecture with exact $D_4$-equivariance guarantees.

\subsection{Datasets}
\label{sec:datasets}

We evaluate on four tile-level classification datasets using the standard splits, summarized in Table~\ref{tab:datasets}: TCGA-UT is split by patient, MHIST uses the provided patient-disjoint split, and both NCT-CRC-HE variants use the standard split with a fixed test set across runs. For the frozen-backbone heads, hyperparameters are selected by 3-fold stratified cross-validation on the training set, with folds fixed across seeds; three folds keep selection tractable at this scale, and all datasets except MHIST leave tens of thousands of tiles per validation fold. Early stopping uses a separate 10\% validation split carved from the training data; the test split is never used during training.

TCGA-UT~\citep{komura2022universal} is the most challenging benchmark with 31 fine-grained tissue classes across multiple organs. NCT-CRC-HE-100K~\citep{kather2019predicting} and its non-normalized variant isolate the impact of stain normalization versus geometric TTA. MHIST~\citep{wei2021petri} is a clinically challenging binary task requiring the differentiation of benign Hyperplastic Polyps (HP) from precancerous Sessile Serrated Adenomas/Polyps (SSA).

\begin{table}[t]
    \centering
    \caption{Datasets used for evaluation. Train and test sizes are tile counts.}
    \label{tab:datasets}
    \vspace{3pt} 
    \begin{tabular}{@{}lrrrl@{}} 
        \toprule
        Dataset & Classes & Train & Test & Domain \\
        \midrule
        TCGA-UT           & 31 & 190k & 40.9k & Multi-organ \\
        NCT-CRC-HE-100K   & 9  & 100k & 7.2k  & Colorectal \\
        NCT-CRC-HE-NoNorm & 9  & 100k & 7.2k  & Colorectal \\
        MHIST             & 2  & 2{,}175 & 977 & Colorectal polyp \\
        \bottomrule
    \end{tabular}
\end{table}

\subsection{TTA Strategies}

We evaluate three TTA strategies of increasing group size:

\begin{enumerate}
  \item \textbf{None}: standard single-view inference (baseline)
  \item \textbf{$V_4$} (4 views): horizontal flip, vertical flip, and both, forming the Klein four-group $V_4 \subset D_4$
  \item \textbf{{$D_4$}} (8 views): all eight elements of the dihedral group, comprising four rotations (0\textdegree, 90\textdegree, 180\textdegree, 270\textdegree) composed with identity and horizontal flip
\end{enumerate}

For each strategy, we evaluate four aggregation methods. Let $G$ denote the set of test-time views, with per-view logits $\ell_g\in\mathbb{R}^C$ and probabilities $p_g=\mathrm{softmax}(\ell_g)$. All sums below are over the group elements $g\in G$
\begin{equation*}
\begin{aligned}
  a &= \mathrm{softmax}\!\left(\frac{1}{|G|}\sum_{g} \ell_g\right) && \text{(logit-mean)}\\
  a &= \frac{1}{|G|}\sum_{g} p_g && \text{(prob-mean)}\\
  a &= \frac{\sum_{g} w_g p_g}{\sum_{g} w_g}, \quad w_g = \max_{c}(p_g)_c && \text{(conf.-weighted)}\\
  \hat{y} &= \arg\max_{c}\, a_c && \text{(prediction)}
\end{aligned}
\end{equation*}
Majority vote predicts $\hat{y}=\operatorname*{mode}_{g}\bigl(\arg\max_{c}\,(p_g)_c\bigr)$.
Unless stated otherwise, we use logit-mean aggregation, which applies Equation~\eqref{eq:reynolds} directly to the logits, as the default; probability-mean, gives the same average gain (within $0.01$ pp). Results for all aggregation variants are reported in Section~\ref{sec:results}.

Finally, we evaluate \textbf{selective TTA}, which adaptively applies full $D_4$ only to uncertain samples based on the entropy of the base-view prediction, trading off accuracy and inference cost (Section~\ref{sec:selective}).

\subsection{Metrics}
\label{sec:metrics}
We report \textbf{balanced accuracy} (macro-averaged recall) as our primary metric, since it weights all classes equally regardless of prevalence, which is critical for the imbalanced TCGA-UT dataset. For a $C$-class problem, balanced accuracy is
\begin{equation*}
  \text{BalAcc}(f)=\frac{1}{C}\sum_{c=1}^{C}\frac{\text{TP}_c}{\text{TP}_c+\text{FN}_c}.
\end{equation*}
The TTA benefit is quantified as the difference in balanced accuracy between TTA and baseline, $\Delta_{\text{TTA}} = \text{BalAcc}(f_{\text{TTA}}) - \text{BalAcc}(f_{\text{base}})$, reported in percentage points (pp).

For each TTA evaluation, we count \emph{corrections}: samples the baseline gets wrong but TTA gets right, and \emph{corruptions}: samples the baseline gets right but TTA gets wrong. The \textbf{correction ratio} $n_{\text{corrected}}/n_{\text{corrupted}} > 1$ indicates net benefit; it measures the \emph{quality} of TTA's prediction flips independently of their total count.

\paragraph{Prediction entropy.} As an uncertainty proxy for selective TTA, we use the Shannon entropy of the single-view softmax $p(y\mid x)\in[0,1]^C$:
\begin{equation*}
  H(x) = -\sum_{c=1}^{C} p(y=c\mid x)\,\log p(y=c\mid x),
\end{equation*}
where $C$ is the number of classes and $\log$ is the natural logarithm. When comparing across datasets with different $C$, we report the normalized entropy $\tilde{H}(x)=H(x)/\log C\in[0,1]$.

\paragraph{Orbit tightness (MPCS).} To quantify representation-level $D_4$ invariance (Section~\ref{sec:orbit}), we ask how close a patch's eight $D_4$-view embeddings $z_g$ are to one another. Let $z_g$ be the frozen-backbone feature vector of view $g$. The mean pairwise cosine similarity (MPCS) averages, over all $\binom{8}{2}$ pairs of views, their cosine similarity,
\begin{equation} \label{MPCS}
  \text{MPCS} = \frac{1}{\binom{8}{2}} \sum_{g < g'} \frac{\langle z_g,\, z_{g'} \rangle}{\lVert z_g\rVert\,\lVert z_{g'}\rVert},
\end{equation}
which equals $1$ when all eight views point in the same direction (a perfectly invariant orbit).

\paragraph{Boundary geometry (margin and straddle ratio).} Whether the orbit's movement changes the prediction depends not on its size but on its position relative to the decision boundary. Let $W\in\mathbb{R}^{C\times D}$ and $\beta\in\mathbb{R}^{C}$ be the linear probe's weights and biases, and let $i,j$ index the top-two classes of the base (identity) view $z_{\mathrm{id}}$. The $i\,|\,j$ boundary is the hyperplane with normal $w_i-w_j$, and the \emph{geometric margin} is the signed distance of the base view from it,
\begin{equation} \label{eq:margin}
  m = \frac{\langle w_i - w_j,\ z_{\mathrm{id}}\rangle + (\beta_i - \beta_j)}{\lVert w_i - w_j \rVert}.
\end{equation}
Let $\hat{n}=(w_i-w_j)/\lVert w_i-w_j\rVert$ be the unit normal to that boundary, the only direction along which motion can change which class wins. The orbit's \emph{spread} along it, and the \emph{straddle ratio}, are
\begin{equation} \label{eq:straddle}
  s=\operatorname{std}_{g\in G}\,\langle z_g,\hat{n}\rangle,
  \qquad
  r = m/s .
\end{equation}
The orbit straddles the boundary when $|r|<1$. To predict flips we score patches by $-|m|$ and $-|r|$, so larger scores indicate more likely flips.

\paragraph{Calibration.} Let $\hat{p}_i=\max_c p(y=c\mid x_i)$ be the confidence of prediction $\hat{y}_i$ on sample $i$. Partitioning the $N$ test samples into ten equal-width confidence bins $B_1,\dots,B_{10}$, the expected calibration error is
\begin{equation*}
  \text{ECE}=\sum_{b=1}^{10}\frac{|B_b|}{N}\,\bigl|\operatorname{acc}(B_b)-\operatorname{conf}(B_b)\bigr|,
\end{equation*}
the sample-weighted gap between accuracy and mean confidence within each bin. Alongside ECE, which is computed from the same top-label confidences, we report the top-label Brier score
\begin{equation*}
  \text{Brier}=\frac{1}{N}\sum_{i=1}^{N}\bigl(\hat{p}_i-\mathbf{1}[\hat{y}_i=y_i]\bigr)^2 .
\end{equation*}

\subsection{Training Protocol}
\label{sec:training}

\paragraph{Frozen backbones and trained heads.} We pre-extract patch embeddings once from each frozen backbone and train lightweight heads on top of the fixed features. Our primary head is a \emph{linear probe}: a single fully-connected layer trained by stochastic gradient descent (Adam) with cross-entropy loss, following the standard self-supervised linear-probe protocol~\citep{oquab2023dinov2}. We additionally report one- and two-hidden-layer MLP probes and $k$-nearest-neighbor baselines ($k\in\{5,10\}$). For the linear and MLP heads, the learning rate ($\{10^{-3},\,3\times10^{-4}\}$) and weight decay ($\{10^{-4},\,10^{-3}\}$) are selected via 3-fold cross-validation on the training set; the head is then trained on the full training set for up to 30 epochs with early stopping (patience 3) on a held-out 10\% validation split, with batch size 512. Hyperparameters are selected once per (model, dataset, head) and shared across seeds.

\paragraph{Finetuned general-purpose models.} The full network is optimized end-to-end with standard augmentation (random $D_4$ transforms; color jitter [brightness, contrast, saturation $\pm 0.25$; hue $\pm 0.05$]; random resized crops [scale $0.6$--$1.0$]; Gaussian blur), differential learning rates ($10^{-4}$ backbone, $10^{-3}$ head), cosine annealing over up to 30 epochs, batch size 64, early stopping patience 5, and label smoothing 0.1.

\paragraph{Seeds.} We run 5 random seeds ($\{0,1,2,3,42\}$) and report mean $\pm$ SEM (standard error of the mean); all patch-level results are averaged over the five seeds; the $k$NN classifier head and all TTA evaluation are deterministic. The seed controls the initialization of the trained head, the mini-batch order, and which 10\% of the training data is held out for early stopping, and, for finetuned models, augmentation sampling. Hyperparameter selection, the frozen embeddings, the test split and TTA evaluation are identical across seeds, so the reported variability reflects training stochasticity alone.

\subsection{Implementation}
\label{sec:implementation}

Repository identifiers, pinned revisions, access conditions and references for all pretrained backbones are listed in Table~\ref{tab:provenance}. D4WRN was constructed using the escnn library~\citep{cesa2022program,weiler2019general} as a Wide ResNet-28-6 with steerable convolutions over the dihedral group $D_4$, yielding $12$M parameters, comparable to ResNet-18---the weakest model. Feature maps use regular field types throughout, aggregating in the final stage to produce $D_4$-invariant embeddings. For the frozen-backbone experiments we pre-extract embeddings for all eight $D_4$ views of every test patch (and a single view of each training patch) in one batched, mixed-precision (FP16) pass per backbone, then train and evaluate all heads on the cached embeddings; per-strategy TTA results are obtained by aggregating the cached per-view embeddings/logits, avoiding redundant forward passes.

\section{Results}
\label{sec:results}
\begin{table*}[t]
    \centering
    \caption{Summary of $D_4$ TTA effects by dataset and model family (frozen linear probes, probability-mean aggregation). \emph{Base} is single-view balanced accuracy; $\Delta_{\text{TTA}}$ is the mean gain in percentage points from $V_4$ and $D_4$ TTA, with the 95\% confidence interval over models for $D_4$. \emph{Corr./Corrupt.} is the ratio of predictions TTA fixes to those it breaks, and \emph{Agree.} the base-vs-$D_4$ prediction agreement rate. General-purpose backbones start lower, agree less across views, and gain more; histology FMs gain little beyond flips.}
    \label{tab:summary}
    \vspace{3pt}
    \footnotesize
    \setlength{\tabcolsep}{3.5pt}
    \begin{tabular}{@{}llrrrrrr@{}}
        \toprule
        & & & \multicolumn{3}{c}{$\Delta_{\text{TTA}}$ (pp)} & & \\
        \cmidrule(lr){4-6}
        Dataset & Family & Base (\%) & $V_4$ & $D_4$ & 95\% CI ($D_4$) & Corr./Corrupt. & Agree. (\%) \\
        \midrule
        TCGA-UT           & Histology FM & $73.8$ & $+0.37$ & $+0.50$ & $[+0.34,+0.66]$ & $1.42$ & $95.6$ \\
                          & General      & $46.4$ & $+1.48$ & $+1.98$ & $[+1.21,+2.76]$ & $1.74$ & $83.0$ \\
        \addlinespace[2pt]
        NCT-CRC-HE-100K   & Histology FM & $94.0$ & $+0.21$ & $+0.29$ & $[-0.14,+0.71]$ & $2.15$ & $98.7$ \\
                          & General      & $89.1$ & $+0.81$ & $+0.93$ & $[+0.50,+1.37]$ & $1.66$ & $96.2$ \\
        \addlinespace[2pt]
        NCT-CRC-HE-NoNorm & Histology FM & $86.9$ & $+0.30$ & $+0.30$ & $[-0.21,+0.82]$ & $1.38$ & $97.2$ \\
                          & General      & $74.3$ & $+0.55$ & $+0.64$ & $[+0.15,+1.12]$ & $1.35$ & $93.1$ \\
        \addlinespace[2pt]
        MHIST             & Histology FM & $77.3$ & $+0.14$ & $+0.11$ & $[-0.23,+0.46]$ & $1.06$ & $96.5$ \\
                          & General      & $76.1$ & $+0.91$ & $+1.56$ & $[+0.65,+2.47]$ & $1.48$ & $92.7$ \\
        \bottomrule
    \end{tabular}
\end{table*}
\subsection{TTA Consistently Improves Predictions}
\label{sec:tta_improves}
Across the 15 frozen-backbone models and 4 datasets, $D_4$ TTA improves balanced accuracy in 50 of 60 model-dataset configurations, indicating that group-averaging is a reliable, but modest, post hoc gain (binomial test vs.\ 50\%: $p < 10^{-6}$). Improvements are larger for general-purpose backbones than for histology FMs (Figure~\ref{fig:tta_overview}). We also observe a clear trend in $D_4$ \emph{prediction stability}: models with smaller $\Delta_{\text{TTA}}$ tend to show lower view-induced flip rates (Figure~\ref{fig:entropy}(a)).

Averaged across all four datasets and 5 seeds, histology FMs gain $+0.30$ pp (95\% CI $[+0.13,+0.47]$ across the 36 model--dataset cells; Wilcoxon signed-rank $p < 10^{-3}$), $4.3\times$ smaller than general-purpose models ($+1.28$ pp; 95\% CI $[+0.94,+1.62]$, $p < 10^{-6}$). The full per-model, per-dataset breakdown is reported in Table~\ref{tab:full_results}. Decomposing each $\Delta_{\text{TTA}}$ into the patches TTA \emph{corrects} versus \emph{corrupts} (Figure~\ref{fig:entropy}(a)), histology FMs show both fewer corrections and a lower net correction rate than general-purpose backbones.

General-purpose backbones gain more than histology FMs on every dataset individually, significantly on all but NCT-CRC-HE-NoNorm (Table~\ref{tab:summary}). The performance gap between model families is significant by a rank-based test that makes no linearity assumption (general $>$ FM, Mann--Whitney U/Wilcoxon rank-sum test $p < 10^{-3}$ on TCGA-UT), and the association between $\Delta_{\text{TTA}}$ and baseline accuracy is negative in all four datasets, strongly on TCGA-UT and NCT-CRC-HE-100K (Spearman rank correlation $\rho = -0.90$, $-0.75$) and weakly on the other two (Fisher's combined $p < 10^{-5}$).
\paragraph{Where the losses occur.} Histology FMs gain on average on every dataset (significantly on TCGA-UT and pooled across datasets) and improve in 27 of their 36 model--dataset combinations (9 FMs $\times$ 4 datasets). The nine losses are small, averaging $-0.21$ pp against $+0.47$ pp for the gains, and concentrate on MHIST, the smallest test set; none occurs on TCGA-UT. Across all models and datasets, TTA corrects 1.6 predictions for every one it corrupts, and every model corrects more than it corrupts (Figure~\ref{fig:entropy}(a)), even where its net gain is small.
\paragraph{Robustness checks.} These conclusions are robust to the choice of lightweight classifier head on top of frozen features (linear vs.\ MLP vs.\ $k$NN), with head-family averages summarized in Table~\ref{tab:head_comparison}. They are likewise robust to pretraining-corpus overlap: excluding the three histology FMs whose pretraining corpora include TCGA slides (Table~\ref{tab:models}) widens the gap on TCGA-UT from $3.9\times$ to $4.7\times$, and the gap replicates on the three patch datasets with no TCGA content.

\paragraph{Headroom.} We test whether the decline in TTA benefit for stronger models is simply due to reduced error (less ``headroom'') by measuring the net correction rate per dataset. On TCGA-UT, the corrected-to-corrupted ratio becomes less favorable as model quality increases (roughly from $\sim\!2{:}1$ for the weakest backbones to $\sim\!1.3{:}1$ for the strongest FMs), suggesting that the remaining mistakes of stronger models are less often fixable by simple geometric averaging.

The remaining gains concentrate among high-entropy predictions (Figure~\ref{fig:entropy}(b--c)), a pattern we explain in Section~\ref{sec:orbit}. Table~\ref{tab:full_results} reports the per-model results for all datasets; on TCGA-UT, general-purpose backbones tend to benefit the most, whereas histology FMs, especially the largest ones, show consistently smaller improvements. 

\paragraph{D4WRN as a near-zero reference.}
\label{sec:d4wrn_caveats}

The D4WRN, exactly $D_4$-equivariant by construction, shows no TTA benefit on average ($+0.00$ pp over all four datasets, within $\pm0.2$ pp on each; Table~\ref{tab:full_results}), despite being much smaller (${\sim}12$M parameters) and having no large-scale pretraining. We treat it as an informative but uncontrolled reference, since it differs from histology FMs in architecture (Wide ResNet vs.\ ViT), scale (86M--1.1B parameters), and training protocol.

\subsection{Frozen Backbones Benefit More Than Finetuned}
\label{sec:frozen_ft}

We evaluate the six general-purpose backbones in both frozen (linear-probe) and finetuned (end-to-end) configurations across all four datasets. Frozen backbones derive a significantly larger benefit from $D_4$ TTA, gaining $+1.28$ pp versus $+0.80$ pp when finetuned---a $1.6\times$ reduction in TTA utility. Importantly, finetuning simultaneously raises the mean baseline balanced accuracy from $71.5\%$ to $79.9\%$ ($+8.4$ pp), meaning this shrinking TTA benefit accompanies a much stronger base model overall.

The reduction varies by dataset: it is largest on the two large, stain-normalized training sets (TCGA-UT $2.5\times$, NCT-CRC-HE-100K $2.3\times$), weaker on NCT-CRC-HE-NoNorm ($1.3\times$), where stain rather than geometric variation dominates the error (Section~\ref{sec:stain}), and absent on MHIST ($1.0\times$), whose $2{,}175$ training patches are too few to instill invariance. Finetuning with augmentation thus absorbs part of the TTA benefit only when training data are sufficient and geometric variation limits accuracy. We conjecture that the same mechanism underlies the small gains of the histology FMs, pretrained with geometric augmentation at far larger scale, but do not test this directly (Section~\ref{sec:limitations}).

\subsection{Strategy Progression: Diminishing Returns Beyond Flips}
\label{sec:strategy}

Doubling the views from $V_4$ to $D_4$ adds little: most of the gain comes from flips alone, especially for histology FMs, for which the proper rotations add almost nothing (Tables~\ref{tab:summary} and~\ref{tab:full_results}). The aggregation method matters little as well: logit-mean and probability-mean averaging are equivalent and confidence weighting is slightly worse, whereas majority voting forfeits about 30\% of the gain.

\subsection{TTA Lowers Predicted Confidence}
\label{sec:calibration}

Histology FMs are not better calibrated than general-purpose backbones; on TCGA-UT they are consistently overconfident (ECE $0.058$ vs.\ $0.011$; Figure~\ref{fig:reliability}). $D_4$ averaging lowers predicted confidence on average, so its effect depends on the starting point: it reduces overconfidence in histology FMs (ECE improves in 26 of 36 model--dataset pairs, Wilcoxon $p = 0.02$), but pushes already well-calibrated general-purpose backbones into underconfidence about as often as it improves them. Pooled over all models, ECE is unchanged, while the top-label Brier score improves ($p < 10^{-8}$), partly reflecting the accuracy gain. TTA thus reduces overconfidence where it is present but does not replace explicit recalibration.

\subsection{Agreement Rate Quantifies Prediction-Level Invariance}
\label{sec:agreement}

Histology FMs achieve a mean base-vs-$D_4$ prediction agreement rate of 97.0\% (range 91.8\%--99.7\%) compared to 91.2\% (range 74.3\%--97.5\%) for general-purpose models; per-dataset values are given in Table~\ref{tab:summary}. This gap quantifies $D_4$ stability at the \emph{prediction} level, which, as Section~\ref{sec:orbit} shows, is a property of the classifier's decision geometry and is \emph{not} explained by embedding-space (representation) invariance. Within the FM family the ranking tracks model scale: CTransPath shows the lowest agreement (91.8\% on TCGA-UT) while the largest models (Virchow-v2, H-Optimus-1) exceed 98\% averaged over datasets.

Baseline softmax entropy also predicts prediction correctness, not just TTA flip events. Across model $\times$ dataset combinations, entropy achieves an Area Under the Receiver Operating Characteristic (AUROC) of $0.849 \pm 0.058$ for discriminating incorrect from correct predictions in histology FMs (general-purpose: $0.816 \pm 0.049$). Incorrect predictions carry $4.3\times$ higher normalized entropy than correct ones for FMs ($2.6\times$ for general-purpose models): when the model is uncertain, it is substantially more likely to be wrong.

\subsection{A Flip Occurs When the Orbit Straddles the Decision Boundary}
\label{sec:orbit}

\begin{figure*}[!t]
    \centering
    \includegraphics[width=0.85\linewidth]{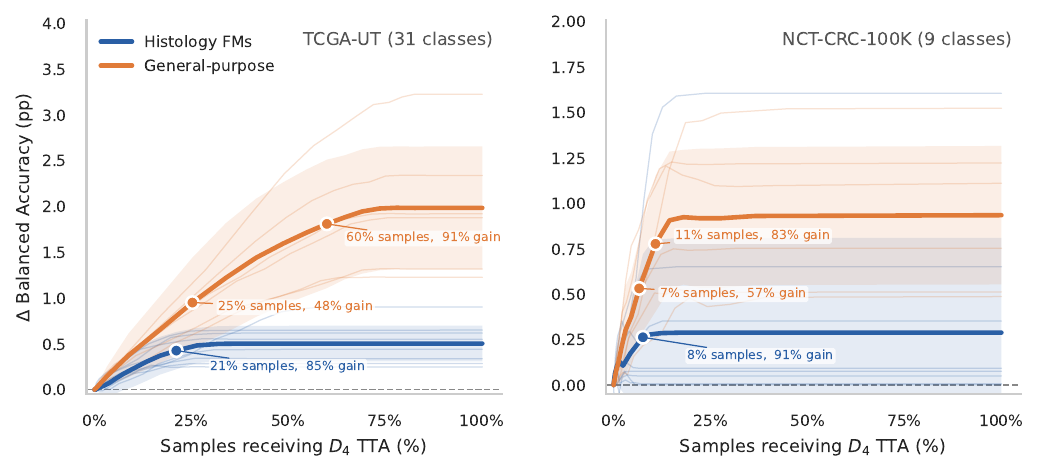}
    \caption{Selective TTA accuracy--efficiency trade-off on TCGA-UT. Each point corresponds to an entropy threshold $t$ for routing samples to full $D_4$ TTA; lower $t$ applies TTA to more samples (higher cost) and vice versa. Histology FMs recover nearly the full $D_4$ benefit while saving a substantial fraction of forward passes. Thin lines indicate individual model curves, the thick line indicates their mean, and the shaded area indicates the confidence interval.}
    \label{fig:selective_pareto}
\end{figure*}

\begin{table}[!tb]
    \centering
    \caption{AUROC for predicting whether $D_4$ TTA flips a patch prediction. Computed per patch for the 15 frozen linear probes on all four patch datasets (mean $\pm$ SD over the 60 model--dataset combinations). Where the orbit sits relative to the decision boundary predicts flips; how large the orbit is does not.}
    \label{tab:flip_predictors}
    \vspace{3pt}
    \begin{tabular}{@{}lccc@{}}
        \toprule
        Predictor & All & Histo FM & General \\
        \midrule
        \multicolumn{4}{@{}l}{\textit{Orbit position relative to the boundary}} \\
        Straddle ratio $|r|$        & $0.954 \pm 0.035$ & $0.974$ & $0.924$ \\
        Geometric margin $|m|$      & $0.947 \pm 0.037$ & $0.969$ & $0.914$ \\
        Softmax entropy (proxy)     & $0.925 \pm 0.054$ & $0.950$ & $0.887$ \\
        \addlinespace[3pt]
        \multicolumn{4}{@{}l}{\textit{Orbit size}} \\
        Orbit spread $s$            & $0.586 \pm 0.056$ & $0.600$ & $0.566$ \\
        Orbit tightness (MPCS)      & $0.568 \pm 0.077$ & $0.570$ & $0.566$ \\
        \bottomrule
    \end{tabular}
\end{table}

We first test the natural hypothesis that representation-level $D_4$ invariance explains TTA benefit, measuring orbit tightness by MPCS \eqref{MPCS}. It does not: histology FMs are about as orbit-invariant as general-purpose models (mean MPCS $0.953$ vs.\ $0.942$), and per patch, orbit tightness barely separates flipped from unflipped predictions (AUROC $0.57$, near chance). The architecturally $D_4$-equivariant D4WRN is the one exception, collapsing each orbit almost to a point (MPCS $\approx 0.999$). This confirms that MPCS detects invariance when it is present, and that FMs' high prediction agreement (Section~\ref{sec:agreement}) is \emph{not} embedding-level invariance.

A flip happens only when the orbit straddles the decision boundary between the base-view top-two classes. The two boundary-geometry quantities, the geometric margin \eqref{eq:margin} and the straddle ratio \eqref{eq:straddle}, separate flipped from unflipped patches almost perfectly and match or exceed softmax entropy, whereas the two orbit-size quantities barely do (Table~\ref{tab:flip_predictors}). Flip risk therefore varies with margin rather than orbit size, and the separation is wider for histology FMs than for general-purpose backbones.

This recasts the FM/general gap as a margin effect, not an invariance one. Since orbit size is essentially equal across families, FMs flip less only because their more separable representations place the same-sized orbit farther from decision boundaries, crossing one $3.4\times$ less often than general models on TCGA-UT (patches with $|r|<1$: $9.9\%$ vs.\ $33.9\%$). Because softmax entropy is a cheap proxy for this margin, it provides an effective signal for deciding which patches to augment, which we exploit next.

\subsection{Selective TTA Preserves Benefits at Reduced Inference Cost}
\label{sec:selective}

Histology FM predictions are highly confident (low entropy) on the large majority of patches, so full $D_4$ TTA adds redundant computation for most inputs. We exploit this by applying selective TTA: single-view inference for samples whose base-view softmax entropy falls below a threshold, and full $D_4$ for uncertain samples:
\begin{equation}
  \hat{f}_{\text{sel}}(x) =
  \begin{dcases}
    f(x) & \text{if } H[\sigma(f(x))] \le t \cdot \ln C \\
    \frac{1}{|G|} \sum_{g \in G} f(g \cdot x) & \text{otherwise},
  \end{dcases}
  \label{eq:selective}
\end{equation}
where normalizing by $\ln C$ makes the threshold $t \in [0,1]$ comparable across datasets with different numbers of classes $C$. Figure~\ref{fig:selective_pareto} shows the efficiency--accuracy tradeoff curve (Pareto frontier) for histology FMs and general-purpose models on TCGA-UT. Averaged over the histology FMs on TCGA-UT:
\begin{itemize}
  \item At $t = 0.25$: only 21\% of samples receive TTA (saving 79\% of forward passes), recovering $+0.43$ pp, about 85\% of the full benefit as in Figure \ref{fig:selective_pareto}.
  \item Moreover, at $t = 0.05$: 46\% of samples receive TTA, recovering 100.0\% of full TTA gain at $+0.50$ pp.
\end{itemize}

For general-purpose models, savings are more modest as their higher disagreement means more samples genuinely benefit. Base-view entropy is thus an effective routing signal, concentrating TTA cost on the few unstable predictions. Routing can concentrate the flips but not improve their quality: among patches that flip, no base-view signal (entropy or margin) separates beneficial corrections from harmful corruptions (AUROC $\approx 0.56$, near chance), so the net benefit of selective TTA is bounded by the intrinsic correction-to-corruption ratio of the flipped set.

\subsection{Stain Normalization Matters More Than Geometric TTA}
\label{sec:stain}

Comparing NCT-CRC-HE-100K (stain-normalized) with NCT-CRC-HE-NoNorm reveals that stain normalization has a far larger effect than geometric TTA. For histology FMs, the mean baseline balanced accuracy drops by $7.2$ pp without stain normalization (94.0\% $\to$ 86.9\%), while TTA recovers an average of $+0.3$ pp on both datasets. This suggests that, on these benchmarks, geometric $D_4$ invariance is not the dominant source of error for modern histology FMs, whereas stain and scanner robustness remains a significant open challenge.

While the preceding results treat tiles as independent samples, clinical deployment and many research pipelines operate on WSIs. We therefore ask whether the (modest) tile-level corrections induced by $D_4$ averaging translate into improvements for slide-level prediction.


\section{WSI-Level Analysis}
\label{sec:wsi}

\subsection{Datasets and Setup}
\label{sec:wsi_setup}

\textbf{Camelyon17}~\citep{bandi2018camelyon17} provides 500 H\&E-stained lymph-node WSIs from 5 acquisition centers (100 slides, 20 patients per center); 50 slides carry pixel-level lesion polygon annotations.
Tissue patches are $256\times256$ at $20\times$ magnification, taken from CLAM segmentation. The frozen UNI patch probe (binary tumor/normal) is evaluated leave-one-patient-out over the 50 annotated slides to avoid leakage. For slide-level MIL we use the standard domain-generalization split: train on centers 0--3 and test on the unseen-scanner center 4 (100 slides).

\textbf{PANDA}~\citep{bulten2022artificial} provides prostate biopsy WSIs with pixel-level Gleason-grade annotations.
We binarize annotations into cancer / non-cancer and extract $256\times256$ patches at $20\times$ magnification, yielding 10,353 slides with at least one patch ($3{,}721$ with $\geq1$ cancer patch).

\paragraph{Slide-level TTA and MIL training.}
Because patch embeddings are cached for all eight $D_4$ views, slide-level TTA can be applied at two points, which are not equivalent because the aggregator is non-linear: \emph{feature-level} averages a patch's eight view embeddings before aggregation, while \emph{logit-level} runs the trained MIL head once per view and averages the slide logits. We report feature-level results; logit-level gives the same null. The MIL head is gated-attention Attention-based Deep Multiple Instance Learning (ABMIL)~\citep{ilse2018abmil} (attention dimension $256$, classifier hidden dimension $512$, dropout $0.25$), trained under cross-entropy with Adam (learning rate $10^{-4}$, weight decay $10^{-5}$), cosine annealing over 50 epochs and early stopping with patience 15. Heads are trained on single-view features and TTA is applied only at inference, so each comparison isolates test-time averaging on a fixed model.

\subsection{Patch Corrections Do Not Transfer to Slide-Level MIL}
\label{sec:wsi_mil}

Adding $D_4$ TTA to a trained ABMIL model~\citep{ilse2018abmil} does not measurably change bag-level performance. On Camelyon17 this null holds across six backbones spanning weak general-purpose models to strong FMs (ResNet-50, ConvNeXtV2, Phikon-v2, GigaPath, UNI and Virchow-v2). Because TTA is toggled at inference on a fixed trained head, each run forms a paired comparison, which bounds the slide-level effect to within $\pm 0.003$ in test AUROC (95\% CI, pooled over backbones and seeds). The design detects changes several times smaller than the gains TTA produces at the patch level, so this bounds the effect rather than merely failing to find one. No backbone shows a change surviving correction for six comparisons. On PANDA, ABMIL quadratic-weighted $\kappa$ is $0.901$ with TTA versus $0.905$ without, when using the UNI backbone evaluated across 3 cross-validation folds.

The null is structural rather than a property of any one backbone. It holds even for ResNet-50 and ConvNeXtV2, whose patch features are far from $D_4$-invariant and which gain the most from patch-level TTA (Section~\ref{sec:results}). ABMIL pools hundreds to thousands of patches per slide, and this aggregation already averages out the small, largely canceling per-patch perturbations that $D_4$ introduces, leaving the bag representation essentially unchanged. The effect is \emph{not} that attention ignores the affected patches: using the patch probe as a diagnostic, the ${\sim}0.1\%$ of patches whose prediction $D_4$ would flip fall in the top decile of UNI ABMIL attention. They are simply too few to move the pooled logit.

At the patch level, the corrections are modest and headroom-gated, as on the tile benchmarks. Leave-one-patient-out per-slide Dice improves on 59\% of PANDA cancer slides (mean $+0.66$\,pp; $n=3{,}721$), with the gain concentrated on poorly-segmented slides ($+1.0$\,pp in the lowest baseline-Dice tertile versus $+0.25$\,pp in the highest). The frozen UNI probe already segments tumor well, leaving little headroom for $D_4$ averaging to recover.


\section{Discussion}
\label{sec:discussion}

\subsection{Prediction Stability Reflects Joint Backbone--Classifier Geometry}

Embedding invariance is neither necessary nor sufficient for stable predictions. Histology FMs reach 97\% $D_4$ agreement with orbits as large as those of general-purpose models, because larger margins keep each orbit on one side of the decision boundary; D4WRN reaches stability the other way, by collapsing its orbits. Two implications follow. First, embedding-level invariance measures such as orbit tightness are a poor guide to whether TTA will help; the margin, or its inexpensive proxy, prediction entropy, is the relevant quantity. Second, because stability depends on the classifier as well as the backbone, TTA benefit is a property of the whole pipeline rather than of the FM alone, consistent with its reduction under finetuning (Section~\ref{sec:frozen_ft}).

\subsection{Practical Recommendations}

{
\begin{enumerate}
  \item \textbf{Histology FMs}: the $+0.30$ pp average gain rarely justifies the $8\times$ inference cost, so skip TTA by default; stain normalization (Section~\ref{sec:stain}) is a higher-impact priority.
  \item \textbf{General-purpose backbones}: TTA is more worthwhile ($+1.28$ pp), especially for frozen linear probes, whose linear head cannot learn invariance from training data ($+0.80$ pp when finetuned; Section~\ref{sec:frozen_ft}).
  \item \textbf{High-stakes applications}: where small gains matter at scale, TTA may justify its cost, since it corrects more predictions than it corrupts.
  \item \textbf{If TTA is used}: average probabilities or logits rather than voting, and apply it selectively (Section~\ref{sec:selective}); routing patches whose normalized entropy exceeds $t \approx 0.25$ augments about one in five FM patches while recovering most of the benefit, and $t$ can be lowered if the full gain is required.
  \item \textbf{Calibration}: for histology FMs, TTA modestly reduces overconfidence; where calibrated probabilities matter, explicit recalibration is cheaper.
\end{enumerate}
}

\subsection{Limitations}
\label{sec:limitations}

Our study has several limitations. First, training protocol is a confound when comparing histology FMs (frozen linear probes only) to general-purpose models (frozen and finetuned). Finetuning reduces mean TTA gain from $+1.28$ pp to $+0.80$ pp, so part of the FM vs.\ general-purpose gap may reflect downstream adaptation rather than pretraining alone. This reduction is itself dataset-dependent (Section~\ref{sec:frozen_ft}), and we do not test it on histology FMs directly. Second, the D4WRN comparison is uncontrolled (architecture, scale, and training protocol all differ; Section~\ref{sec:d4wrn_caveats}); a matched equivariant control is left to future work. Third, our WSI analysis (Section~\ref{sec:wsi}) is limited to two datasets with patch-level annotations (Camelyon17, PANDA); generalization to other cancer types and scanners remains to be evaluated. Fourth, our statistical tests (binomial test, Wilcoxon signed-rank, Mann--Whitney/Wilcoxon rank-sum, Spearman correlation with Fisher's combined $p$-value) are each reported without correction for the multiple comparisons made across datasets and models; future work could apply multiple-comparison corrections. Fifth, datasets are limited to H\&E staining; immunohistochemistry or other stain types may have different symmetry properties.

\subsection{Future Directions}

Our results point to several promising research directions:

\begin{itemize}
  \item \textbf{Stain-equivariant architectures}: Analogous to $D_4$-equivariant networks for geometric transforms, architectures with built-in stain invariance could address the larger remaining gap.
  \item \textbf{Adaptive thresholding for selective TTA}: We demonstrate entropy-based routing at fixed thresholds (Section~\ref{sec:selective}); future work could learn per-dataset or per-class thresholds, or jointly optimize the routing policy with the classifier head.

  \item \textbf{Invariance probing}: Developing systematic benchmarks for measuring the degree of learned invariance across different transformation groups (geometric, photometric, stain) would help characterize FM capabilities beyond standard accuracy metrics.
\end{itemize}

\section{Conclusion}
\label{sec:conclusion}

We have presented the first large-scale evaluation of TTA across 16 vision models and 4 histopathology datasets. Our results show that modern histology FMs exhibit high prediction stability under $D_4$ transforms and gain only $+0.30$ pp from $D_4$ TTA, a $4.3\times$ smaller benefit than general-purpose models. While TTA remains a safe and reliable technique, improving most configurations and correcting more predictions than it corrupts, its marginal benefit for histology FMs suggests that geometric $D_4$ robustness is often not the primary performance bottleneck for tile-level classification with modern FMs.

We further show that entropy-based selective TTA, which routes only uncertain samples to full $D_4$, recovers over half of the TTA benefit for histology FMs while augmenting only ${\sim}13\%$ of patches, providing a practical path to reduced-cost TTA when inference budget is constrained. By contrast, stain normalization has a far larger effect on these models, pointing to stain and scanner robustness as the more pressing open challenge. We release our complete evaluation framework to facilitate further research into invariance properties of histopathology foundation models.


\acks{We utilized generative artificial intelligence (AI) to assist in creating and refining code segments based on specified tasks, including debugging, editing, and autocompleting code. Additionally, generative AI tools were used to help
structure sentences and perform grammar checks. The conceptualization and ideation, as well as all instructions given to the AI, were exclusively the result of the authors’
creative and intellectual input. We take full responsibility for reviewing and ensuring the accuracy and quality of all AI-generated content included in this work.}

\data{All datasets are available from their original providers: TCGA-UT~\citep{komura2022universal} (Hugging Face \nolinkurl{dakomura/tcga-ut}); NCT-CRC-HE-100K, its non-normalized variant and CRC-VAL-HE-7K~\citep{kather2019predicting} (Zenodo; we use the Hugging Face mirror \nolinkurl{1aurent/NCT-CRC-HE}); MHIST~\citep{wei2021petri} (from the dataset authors upon completion of a data-use form); Camelyon17~\citep{bandi2018camelyon17} (Camelyon17 challenge); and PANDA~\citep{bulten2022artificial} (Kaggle, under the competition terms). Repositories, revisions and access conditions of all pretrained backbones are listed in Table~\ref{tab:provenance}. Code for embedding extraction, training, TTA evaluation and all analyses, together with configuration files, the seeded procedure that regenerates the PANDA cross-validation folds, and the compiled results underlying all tables and figures, is available at \url{https://github.com/lars-chen/histology-tta}. Cached embeddings are not redistributed given the license terms of several source models; they can be regenerated with the released code.}

\bibliographystyle{plainnat}
\bibliography{sample}


\clearpage
\appendix
\onecolumn

\section{Additional Tables and Figures}\suppressfloats[t]

\begin{table*}[t]
\centering
\caption{Per-model TTA results across all four datasets (frozen backbone / SGD linear probe,
probability-mean aggregation, 5 seeds). \textit{Base}: no-TTA balanced accuracy (\%).
$V_4$ and $D_4$: change in percentage points from 4-view (flips) and 8-view (full dihedral)
TTA respectively. Significance of $\Delta_{\text{TTA}}$ ($D_4$) vs.\ zero (two-sided paired $t$-test, $n=5$ seeds,
uncorrected): $^{*}p<0.05$, $^{**}p<0.01$, $^{***}p<0.001$.
Best $\Delta_{\text{TTA}}$ ($D_4$) per dataset in \textbf{bold}. Class counts: TCGA-UT 31, NCT-100K / NCT-NoNorm 9, MHIST 2.
D4WRN is finetuned (seeds vary by dataset; see $^\dagger$) and shown as an
uncontrolled equivariant reference (see Section~\ref{sec:d4wrn_caveats}).}
\label{tab:full_results}
\vspace{4pt}
\footnotesize
\setlength{\tabcolsep}{2.5pt}
\resizebox{\linewidth}{!}{%
\begin{tabular}{@{}llcccccccccccc@{}}
\toprule
& & \multicolumn{3}{c}{TCGA-UT} & \multicolumn{3}{c}{NCT-100K} & \multicolumn{3}{c}{NCT-NoNorm} & \multicolumn{3}{c}{MHIST} \\
\cmidrule(lr){3-5} \cmidrule(lr){6-8} \cmidrule(lr){9-11} \cmidrule(lr){12-14}
Type & Model & Base & $V_4$ & $D_4$ & Base & $V_4$ & $D_4$ & Base & $V_4$ & $D_4$ & Base & $V_4$ & $D_4$ \\
\midrule
\multicolumn{14}{@{}l}{\textit{General-purpose}} \\
& ResNet-18          & 37.5 & $+2.48$ & \textbf{$+3.23^{***}$} & 86.9 & $+1.06$ & $+1.11^{***}$ & 69.5 & $+0.72$ & $+0.66^{*}$ & 71.8 & $+0.94$ & $+1.84^{***}$ \\
& ResNet-50          & 42.5 & $+1.92$ & $+2.34^{***}$ & 87.4 & $+1.10$ & $+1.52^{***}$ & 80.4 & $+0.85$ & $+0.82^{***}$ & 72.8 & $+1.27$ & $+2.01^{**}$ \\
& ConvNeXtV2-Tiny    & 48.3 & $+1.31$ & $+1.92^{***}$ & 88.8 & $+0.43$ & $+0.49^{**}$ & 70.4 & $+0.47$ & $+1.05^{***}$ & 76.2 & $+1.43$ & \textbf{$+2.48^{**}$} \\
& ConvNeXtV2-Base    & 49.6 & $+1.01$ & $+1.32^{***}$ & 90.1 & $+0.83$ & $+1.22^{***}$ & 69.3 & $+0.20$ & $-0.24^{**}$ & 77.8 & $+0.43$ & $+0.94$ \\
& DINOv2-S           & 49.8 & $+1.27$ & $+1.88^{***}$ & 90.7 & $+0.78$ & $+0.75^{***}$ & 75.5 & $+0.40$ & $+0.94^{**}$ & 77.3 & $+1.28$ & $+1.98^{***}$ \\
& DINOv2-B           & 50.7 & $+0.90$ & $+1.23^{***}$ & 90.7 & $+0.68$ & $+0.51^{**}$ & 80.8 & $+0.64$ & $+0.58^{***}$ & 80.4 & $+0.11$ & $+0.12$ \\
\addlinespace[3pt]
\multicolumn{14}{@{}l}{\textit{Histology foundation models}} \\
& CTransPath         & 57.9 & $+0.67$ & $+0.90^{***}$ & 94.2 & $+0.48$ & $+0.35^{**}$ & 80.2 & $+0.27$ & $+0.00$ & 59.0 & $+0.09$ & $-0.58$ \\
& Phikon             & 74.0 & $+0.45$ & $+0.65^{***}$ & 92.2 & $+0.48$ & $+0.65^{***}$ & 76.4 & $+2.34$ & \textbf{$+1.66^{***}$} & 73.8 & $+0.28$ & $+0.72$ \\
& Phikon-v2          & 76.8 & $+0.31$ & $+0.44^{***}$ & 91.6 & $+0.89$ & \textbf{$+1.60^{***}$} & 89.5 & $-0.14$ & $-0.17$ & 71.2 & $-0.05$ & $-0.16$ \\
& UNI                & 72.9 & $+0.41$ & $+0.55^{***}$ & 93.2 & $-0.26$ & $-0.24^{*}$ & 85.8 & $-0.35$ & $-0.38^{*}$ & 81.2 & $-0.50$ & $-0.33$ \\
& UNI-v2             & 78.5 & $+0.51$ & $+0.62^{***}$ & 95.5 & $-0.04$ & $+0.00$ & 93.0 & $+0.13$ & $+0.01$ & 81.1 & $+0.97$ & $+0.71$ \\
& Virchow            & 73.3 & $+0.46$ & $+0.50^{***}$ & 94.5 & $+0.01$ & $+0.01$ & 81.6 & $-0.17$ & $+0.25$ & 81.6 & $+0.42$ & $+0.48^{*}$ \\
& Virchow-v2         & 77.9 & $+0.20$ & $+0.28^{***}$ & 94.8 & $+0.13$ & $+0.05$ & 91.6 & $+0.14$ & $+0.14$ & 86.7 & $+0.35$ & $+0.20$ \\
& GigaPath           & 73.7 & $+0.24$ & $+0.34^{***}$ & 95.3 & $+0.05$ & $+0.09$ & 89.4 & $+0.38$ & $+1.18^{***}$ & 80.4 & $-0.37$ & $-0.02$ \\
& H-Optimus-1        & 79.4 & $+0.13$ & $+0.25^{**}$ & 95.0 & $+0.12$ & $+0.08$ & 94.4 & $+0.13$ & $+0.07$ & 80.8 & $+0.05$ & $+0.00$ \\
\addlinespace[3pt]
\multicolumn{14}{@{}l}{\textit{Equivariant reference}} \\
& D4WRN$^\dagger$    & 69.6 & $+0.17$ & $+0.17$ & 94.3 & $+0.08$ & $+0.08$ & 72.6 & $-0.19$ & $-0.18$ & 72.3 & $-0.10$ & $-0.06$ \\
\bottomrule
\end{tabular}}
\vspace{4pt}

\noindent\small$^\dagger$ D4WRN is evaluated in finetuned mode; number of seeds varies by dataset
(TCGA-UT: 5, NCT-CRC-100K/MHIST: 2, NCT-CRC-NoNorm: 1). All other models use frozen backbone
with SGD linear probe (5 seeds).
\end{table*}
\begin{table*}[t]
\centering
\caption{Mean $D_4$ TTA benefit ($\Delta_{\text{TTA}}$ in balanced accuracy, pp) by classifier head type and
model family (frozen backbone, probability-mean aggregation). Values averaged over all models
within each family and 5 seeds (kNN: 1 seed, deterministic).
\textit{Base}: mean no-TTA balanced accuracy (\%). $\Delta_{\text{TTA}}$: pp gain from $D_4$ TTA.}
\label{tab:head_comparison}
\vspace{4pt}
\footnotesize
\setlength{\tabcolsep}{3pt}
\begin{tabular}{@{}llcccccccc@{}}
\toprule
& & \multicolumn{2}{c}{TCGA-UT (31 cls)} & \multicolumn{2}{c}{NCT-CRC-100K (9 cls)} & \multicolumn{2}{c}{NCT-CRC-NoNorm (9 cls)} & \multicolumn{2}{c}{MHIST (2 cls)} \\
\cmidrule(lr){3-4} \cmidrule(lr){5-6} \cmidrule(lr){7-8} \cmidrule(lr){9-10}
Family & Head & Base & $\Delta_{\text{TTA}}$ & Base & $\Delta_{\text{TTA}}$ & Base & $\Delta_{\text{TTA}}$ & Base & $\Delta_{\text{TTA}}$ \\
\midrule
\multicolumn{10}{@{}l}{\textit{General-purpose models}} \\
& kNN ($k$=10)       & 27.6 & $+0.86$ & 83.9 & $+0.99$ & 80.3 & $+1.19$ & 71.9 & $+0.95$ \\
& Linear             & 46.4 & $+1.98$ & 89.1 & $+0.93$ & 74.3 & $+0.64$ & 76.1 & $+1.56$ \\
& MLP-1h             & 48.4 & $+2.33$ & 89.8 & $+1.17$ & 73.8 & $+1.06$ & 76.8 & $+1.73$ \\
& MLP-2h             & 48.3 & $+2.38$ & 89.6 & $+1.13$ & 73.5 & $+1.37$ & 77.3 & $+1.67$ \\
\addlinespace[3pt]
\multicolumn{10}{@{}l}{\textit{Histology FMs}} \\
& kNN ($k$=10)       & 63.0 & $+0.32$ & 94.2 & $+0.23$ & 89.9 & $-0.01$ & 73.8 & $-0.14$ \\
& Linear             & 73.8 & $+0.50$ & 94.0 & $+0.29$ & 86.9 & $+0.30$ & 77.3 & $+0.11$ \\
& MLP-1h             & 74.5 & $+0.39$ & 94.0 & $+0.30$ & 86.4 & $+0.39$ & 80.2 & $+0.33$ \\
& MLP-2h             & 74.5 & $+0.40$ & 93.9 & $+0.36$ & 85.5 & $+0.48$ & 80.0 & $+0.45$ \\
\bottomrule
\end{tabular}
\end{table*}
\begin{table}[H]
    \centering
    \caption{Pretrained backbones. \emph{Revision}: commit id, model name, or torchvision weights. \emph{Agreement}: model requires accepting a license agreement.}
    \label{tab:provenance}
    \vspace{3pt}
    \footnotesize
    \setlength{\tabcolsep}{3pt}
    \begin{tabular}{@{}l p{0.36\linewidth} l l l@{}}
        \toprule
        Model & Repository & Revision & Access & Reference \\
        \midrule
        \multicolumn{5}{@{}l}{\textit{Histology foundation models}} \\
        \addlinespace[3pt]
        CTransPath  & \raggedright \href{https://huggingface.co/1aurent/swin_tiny_patch4_window7_224.CTransPath}{\nolinkurl{1aurent/swin_tiny_patch4_window7_224.CTransPath}} & \texttt{59c740e} & open & \citet{wang2022ctranspath} \\
        Phikon      & \raggedright \href{https://huggingface.co/owkin/phikon}{\nolinkurl{owkin/phikon}}               & \texttt{057cc02} & open      & \citet{filiot2023phikon} \\
        Phikon-v2   & \raggedright \href{https://huggingface.co/owkin/phikon-v2}{\nolinkurl{owkin/phikon-v2}}            & \texttt{2ae989a} & open      & \citet{filiot2024phikon2} \\
        UNI         & \raggedright \href{https://huggingface.co/MahmoodLab/UNI}{\nolinkurl{MahmoodLab/UNI}}             & \texttt{b55a5ec} & agreement & \citet{chen2024uni} \\
        UNI-v2      & \raggedright \href{https://huggingface.co/MahmoodLab/UNI2-h}{\nolinkurl{MahmoodLab/UNI2-h}}          & \texttt{d517a8d} & agreement & \citet{chen2024uni} \\
        Virchow     & \raggedright \href{https://huggingface.co/paige-ai/Virchow}{\nolinkurl{paige-ai/Virchow}}           & \texttt{19eebc8} & agreement & \citet{vorontsov2024virchow} \\
        Virchow-v2  & \raggedright \href{https://huggingface.co/paige-ai/Virchow2}{\nolinkurl{paige-ai/Virchow2}}          & \texttt{3158645} & agreement & \citet{zimmermann2024virchow2} \\
        GigaPath    & \raggedright \href{https://huggingface.co/prov-gigapath/prov-gigapath}{\nolinkurl{prov-gigapath/prov-gigapath}} & \texttt{eba85dd} & agreement & \citet{xu2024gigapath} \\
        H-Optimus-1 & \raggedright \href{https://huggingface.co/bioptimus/H-optimus-1}{\nolinkurl{bioptimus/H-optimus-1}}      & \texttt{910efa9} & agreement & \citet{hoptimus2024} \\
        \addlinespace[8pt]
        \multicolumn{5}{@{}l}{\textit{General-purpose models}} \\
        \addlinespace[3pt]
        ConvNeXtV2-T & \raggedright \href{https://huggingface.co/timm/convnextv2_tiny.fcmae_ft_in22k_in1k}{\nolinkurl{timm/convnextv2_tiny.fcmae_ft_in22k_in1k}} & \texttt{b1dd462} & open & \citet{woo2023convnextv2} \\
        ConvNeXtV2-B & \raggedright \href{https://huggingface.co/timm/convnextv2_base.fcmae_ft_in22k_in1k}{\nolinkurl{timm/convnextv2_base.fcmae_ft_in22k_in1k}} & \texttt{9e250ed} & open & \citet{woo2023convnextv2} \\
        DINOv2-S/B   & \raggedright \texttt{torch.hub}: \href{https://github.com/facebookresearch/dinov2}{\nolinkurl{facebookresearch/dinov2}} & \texttt{vits14}/\texttt{vitb14} & open & \citet{oquab2023dinov2} \\
        ResNet-18/50 & \raggedright \texttt{torchvision} & \texttt{V1}/\texttt{V2} & open & \citet{he2016resnet} \\
        \bottomrule
    \end{tabular}
\end{table}

\begin{figure}[H]
    \centering
    \includegraphics[width=0.85\linewidth]{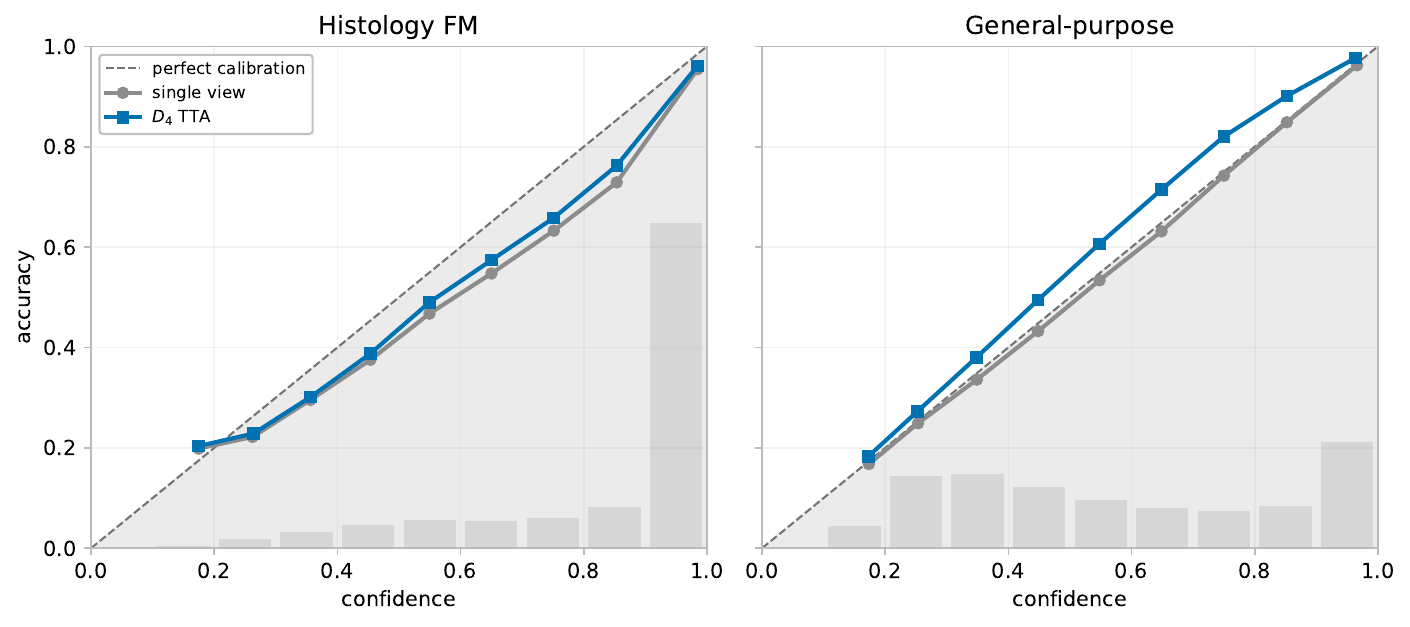}
    \caption{Reliability on TCGA-UT, single-view versus $D_4$ TTA. Points below the diagonal indicate overconfidence; grey bars show the share of predictions in each confidence bin. Curves and ECE are computed per model and seed and averaged within each family; bins holding fewer than $0.5\%$ of a model's predictions are omitted from the curve but retained in the ECE. Histology FMs are overconfident, and $D_4$ TTA moves them toward the diagonal (ECE $0.058 \to 0.049$); general-purpose backbones are near-calibrated, and TTA pushes them into underconfidence ($0.011 \to 0.036$). Histology FMs place $65\%$ of their predictions in the top confidence bin, against $21\%$ for general-purpose backbones.}
    \label{fig:reliability}
\end{figure}

\end{document}